\documentclass[letterpaper, 10 pt, conference]{ieeeconf}  

\IEEEoverridecommandlockouts                              

\usepackage{graphics} 
\usepackage{epsfig}
\usepackage{amsmath}
\usepackage{amssymb}
\usepackage{algorithm}
\usepackage{algorithmicx}
\usepackage{algpseudocode}
\usepackage{booktabs}

\usepackage{flushend}

\usepackage{cite}
\usepackage[caption=false]{subfig}

\usepackage{hyperref}

\newcommand{\figref}[1]{Fig. \ref{#1}}

\usepackage{color}

\usepackage[font=footnotesize,labelfont=bf]{caption}
\title{\LARGE \bf
Learning from Extrapolated Corrections
}

\author{Jason Y. Zhang and Anca D. Dragan
\thanks{Department of Electrical Engineering and Computer
Sciences, University of California, Berkeley, Berkeley 94720}
\thanks{\tt{\{zhang.j,anca\}@berkeley.edu}}%
}

\newcommand{\norm}[1]{\left\lVert#1\right\rVert}
\newcommand{\trans}{^\top}

\begin{document}

\maketitle
\thispagestyle{empty}
\pagestyle{empty}

\begin{abstract}
Our goal is to enable robots to learn cost functions from user guidance. Often it is difficult or impossible for users to provide full demonstrations, so corrections have emerged as an easier guidance channel. However, when robots learn cost functions from corrections rather than demonstrations, they have to extrapolate a small amount of information -- the change of a waypoint along the way -- to the rest of the trajectory. We cast this extrapolation problem as online function approximation, which exposes different ways in which the robot can interpret what trajectory the person intended, depending on the function space used for the approximation. Our simulation results and user study suggest that using function spaces with non-Euclidean norms can better capture what users intend, particularly if environments are uncluttered. This, in turn, can lead to the robot learning a more accurate cost function and improves the user's subjective perceptions of the robot.

\end{abstract}

\section{Introduction}

Robots typically generate their motion to optimize some cost function \cite{chomp,trajopt,jaillet2008,karaman2011}. Specifying good cost functions for robot motion planning is difficult for two reasons. First, \emph{tuning} cost function parameters to get the desired behavior in a single environment, let alone across a range of test environments, can be challenging, as different criteria that are important can be at odds with each other. Second, the designer who is supposed to specify the cost function might actually \emph{not know it}: we design robots to help end-users, and how the end-users want the robot to move is up to them. Different people might have different preferences, e.g. how far away the robot needs to stay from them as it moves, how much it should try to stay in the user's visible space, etc.

Inverse Reinforcement Learning \cite{abbeel2004, ng2000} is an excellent alternative to manually specifying the cost function. The designer or the user provides the robot with demonstrated trajectories, and the robot infers the cost function that explains the demonstrations. This has been successful in many domains, including driving and social navigation \cite{levine2012, ziebart2009, ratliff2006, kretzschmar2014, kretzschmar2016}. It has been applied to manipulation in some settings, but it remains difficult to use as a cost learning tool because demonstrations are difficult to provide in manipulation since users must coordinate many degrees of freedom over time \cite{akgun2012, argall2008}.

As a result, a relatively new line of work focuses on learning from \emph{corrections} rather than full demonstrations \cite{jain2013}. Corrections leverage the idea that while providing a sequence of configurations over time is challenging, people can provide a single configuration easily. Rather than generating a trajectory from scratch, the person can modify an existing trajectory by taking one of its waypoints and physically changing it to a new configuration.
\begin{figure}
    \centering
    \includegraphics[width=\columnwidth]{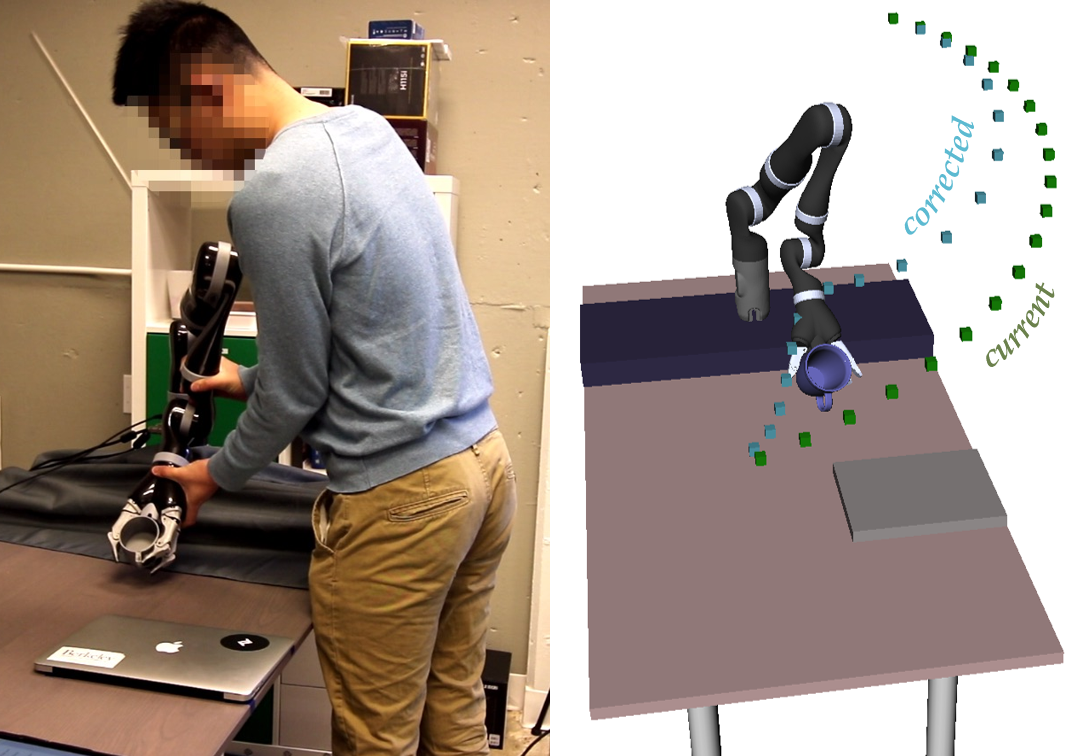}
    \caption{\textbf{Left}: User provides a correction to teach the robot to avoid the laptop. \textbf{Right}: Blue trajectory represents one possible interpretation of the correction to the green trajectory.}
    \label{fig:user}
\end{figure}

With the move from a full demonstration to a corrected waypoint, however, comes a big loss in the amount of information the robot can access. It must now \emph{infer} the entire trajectory from one waypoint. The assumptions we make when performing this extrapolation can affect the quality of the learning.

In other words, the robot has to \emph{estimate} what trajectory the person might have \emph{intended} given its current trajectory and the corrected configuration. Prior work performs this estimation by implicitly assuming that only the one corrected waypoint should change, and the rest of the trajectory should stay the same \cite{jain2013}; or proposes particular ways to deform the trajectory based on the correction \cite{bajcsy2017}.

Building on work that has explored deforming trajectories based on changing waypoints in contexts outside of learning cost functions \cite{dragan2015,losey2017}, we cast the problem of estimating the full trajectory explicitly as a function approximation problem: we have a current estimate (the current trajectory), we receive one new data point (that the corrected timepoint maps to the corrected configuration), and we re-estimate our trajectory online based on this new data point. Different choices of the space of functions we use for approximation (different inner products) map to different assumptions about what the user intended in prior work, with a notable difference between Euclidean \cite{jain2013} and non-Euclidean \cite{bajcsy2017} inner products.

Naturally, when learning cost functions from corrections, we want to understand which choice better matches what users \textit{actually} intend, and, more importantly, which leads to the most effective learning. We analyze these questions in simulation and in a user study and find that non-Euclidean inner products that correlate trajectory waypoints across time can often lead to higher user ratings when it comes to how closely the correction matches the user's intention, in particular for uncluttered scenes. Further, we see that the learning performance is higher with such a choice, both subjectively (as perceived by users) and objectively (when measuring how well the robot learned the desired cost function in a controlled task).  

\noindent\textbf{Summary of Contributions.} Overall, we find that learning cost functions from corrections benefits from explicitly attempting to estimate the trajectory the user might have intended. It is this explicit estimation lens that exposes our choices for how to interpret and extrapolate from corrections, challenging or validating assumptions we've made in the past. From the perspective of work that uses non-Euclidean norms to deform trajectories \cite{losey2017, dragan2015}, we validate that these are also useful when \emph{learning cost functions} based on the deformed trajectories. From the perspective of work that learns cost functions \cite{jain2013, bajcsy2017}, we challenge the notion that Euclidean norms are always best \cite{jain2013}, and support the choice to sometimes use non-Euclidean deformation \cite{bajcsy2017}.

\section{Learning Cost from Corrections}

\noindent\textbf{Problem Statement.} We denote a robot trajectory by $\xi = (q_0,\hdots,q_T)$, which is represented as a sequence of configurations from a start to final time. In any environment, the robot needs to minimize a cost function $\mathcal{U}$ which we parametrize as a linear combination of features \cite{ng2000,shivaswamy2015,jain2013}:

\begin{equation}
	\mathcal{U}(\xi) = w^H \cdot \phi(\xi)
\end{equation}
where $w^H$ is a weight vector and $\phi(\cdot)$ featurizes the trajectory. 

The robot does not observe $w^H$ -- only how an end-user or robot designer might want the robot to move. 
We assume that the user or designer implicitly knows the correct $w^H$ but cannot directly explicate it (end users cannot write down cost parameters, and even designers have trouble tuning them in a generalizable way). Instead, the user can \emph{correct} any current robot trajectory $\xi$ to a  trajectory $\bar{\xi}$ such that
$$w^H\cdot \phi(\bar{\xi})<w^H \cdot \phi(\xi)$$

The robot is penalized according to the ground truth $w^H$ and needs to estimate it from the human's guidance in order to perform well. 

This problem can be characterized as acting in a partially observed system in which at every step the robot executes a trajectory and transitions to a new environment. In this formalism, $w^H$ is the hidden state in the system and the user's corrected trajectories at every step are observations about $w^H$.

\noindent\textbf{Solution.}
Solutions to this problem tend to separate estimating the true weights from finding the best motion plan \cite{jain2013,ratliff2006,bajcsy2017}. At every step $i$, the robot maintains an estimate of $w^H$ (either $w_i$ or a belief $b_i(w)$) and uses it to generate the optimal trajectory $\xi_i$:
$$\xi_i=\arg\min_{\xi}w_i\cdot\phi(\xi) $$
in the case of a running estimate, or 
$$\xi_i=\arg\min_{\xi}\mathbb{E}[w\cdot\phi(\xi)|w \sim b_i]$$
in the case of a belief.

When the human provides a correction, the robot infers a corrected trajectory $\bar{\xi_i}$ and needs to update its estimate of the weight vector. Under an observation model where corrected trajectories are exponentially more probable when they have lower true cost and a Gaussian prior over $w$, \cite{bajcsy2017} has shown that the MAP can be approximated as
$$w_{i+1}=\arg\min_{w} w\cdot \phi(\bar{\xi_i})-w\cdot \phi(\xi_i) + \frac{1}{2\beta} ||w-w_i||^2 $$
for positive $\beta$. This has the intuitive interpretation of finding a cost function in which the corrected trajectory is better than the original trajectory but not deviating too far from the previous estimate.

Taking the gradient and setting it to 0, the new estimate becomes:
$$w_{i+1}=w_i - \beta (\phi(\bar{\xi_i})-\phi(\xi_i)) $$
This is the same update rule used in co-active learning \cite{jain2013} and Online Maximum Margin Planning \cite{ratliff2006} if corrections were demonstrations.

\section{``Intended'' Corrections}
Since the goal of learning from corrections is to address cases where users would have a difficult time demonstrating trajectories, it is important to note that the robot does not actually measure a corrected trajectory $\bar{\xi}$ directly. Instead, after observing the entire trajectory, users correct one point along the path, from $q_t$ to $\bar{q_t}$. Thus, when the robot gets a correction, it does not observe what corrected trajectory the user \emph{intends}, it just observes one point along that intended trajectory and needs to infer the rest.

\subsection{State of the art}
Jain et al. \cite{jain2013} assume that given a trajectory $\xi=(q_0,..,q_t,..,q_T)$ and a correction $\bar{q_t}$, the intended correction at the trajectory level is $\bar{\xi}=(q_0,..,\bar{q_t},..,q_T)$, i.e. that the user only meant to change a single waypoint along the trajectory.

While this $\bar{\xi}$ might make sense in the context of RRT \cite{lavalle2000} trajectories which have a few waypoints that are far apart and require large and almost instantaneous changes in velocity to follow, it is likely not what users have in mind when correcting smooth robot trajectories made out of many waypoints (as is typical of trajectories produced via trajectory optimization \cite{chomp,trajopt}). If the original trajectory $\xi$ is smooth, then $\bar{\xi}$ will jerk when going from $q_{t-1}$ to $\bar{q_t}$ to $q_{t+1}$. 
Not only is this likely not what user intends, but it might also fail to substantially improve the original trajectory, especially if smoothness is part of the ground truth cost.

Bajcsy et al. \cite{bajcsy2017} assume that the intended correction is a deformation of the original trajectory, propagating the change at one waypoint down to the rest of the configurations by multiplying through a linear operator. 

In what follows, we generalize this assumption. We present a formalism for deriving the corrected trajectory based on prior work in Dynamic Movement Primitives \cite{dragan2015} and physical human-robot interaction \cite{losey2017}, and show how the assumption above is one instance of an entire family of possible interpretations for what the user intended.

\subsection{Formalism of intended corrections}

The only thing that the robot knows about the intended correction $\bar{\xi}$ is that it should go through the corrected waypoint $\bar{q_t}$, meaning $\bar{\xi}(t)=\bar{q_t}$. We treat finding $\bar{\xi}$ as one step of an online function approximation problem -- we are at $\xi$, have received a data point $(t,\bar{q_t})$, and want to minimally update our estimate to incorporate this new data point:

\begin{equation}
	\begin{aligned}
	& \underset{\bar\xi}{\min}
	& & \frac{1}{2}\norm{\bar\xi-\xi}^2_A \\
	& \text{s.t.}
	& & \bar\xi(t) = \bar{q_t} \\
	& & & \bar\xi(0) = \xi(0) \\
	& & & \bar\xi(T) = \xi(T)
	\end{aligned}
	\label{eqn:optimization}
\end{equation}

Here, distance to the original trajectory $\xi$ is measured with respect to some inner product $A$ in the Hilbert space of trajectories. 

\noindent\textbf{Solution:} If the trajectory is discretized by time, then $A$ is a matrix, so $\norm{\xi}_A^2 = \xi\trans A \xi$. The Lagrangian of \eqref{eqn:optimization} is
\begin{align*}
    \mathcal{L} = \frac{1}{2}(\bar{\xi} - \xi)\trans A (\bar{\xi} - \xi) + \lambda (\bar\xi(t) - \bar{q_t}) \\
    + \gamma (\bar\xi (0) - \xi(0)) + \kappa (\bar\xi(T) - \xi(T))
\end{align*}
Set the gradients w.r.t. $\bar\xi$, $\lambda$, $\gamma$, and $\kappa$ to 0:
\begin{equation}
\begin{split}
    \nabla_{\bar\xi}\mathcal{L} = A(\bar\xi - \xi) + \begin{bmatrix}\gamma, 0, \ldots, 0, \lambda, 0, \ldots, 0, \kappa \end{bmatrix}\trans = 0\\
        \nabla_\lambda \mathcal{L} = \bar\xi(t) - \bar{q_t} = 0 \\
        \nabla_\gamma \mathcal{L} = \bar\xi(0) - \xi(0) = 0 \\
    \nabla_\kappa \mathcal{L} = \bar\xi(T) - \xi(T) = 0\\
\end{split}
\label{eqn:gradients}
\end{equation}
Thus, the solution to \eqref{eqn:optimization} is:
\begin{equation}
    \bar\xi = \xi - A^{-1}\begin{bmatrix}\gamma, 0, \ldots, 0, \lambda, 0, \ldots, 0, \kappa \end{bmatrix}\trans
\end{equation}
where $\lambda$, $\gamma$, and $\kappa$ satisfy \eqref{eqn:gradients}.

This has an intuitive interpretation: the robot uses the inverse of the norm to \emph{propagate} the correction to the rest of the trajectory, while keeping the start and the goal fixed. This is analogous to using a norm to respond to changes in the goal, as in \cite{dragan2015}, and similar to work in responding to a force during haptic robot teleoperation \cite{losey2017} (there, the propagation happens not from the current point, but from a future time point, so that the human does not have to keep providing input).

\subsection{Metrics for interpreting corrections}

\begin{figure}
		\centering
		\includegraphics[width=\columnwidth]{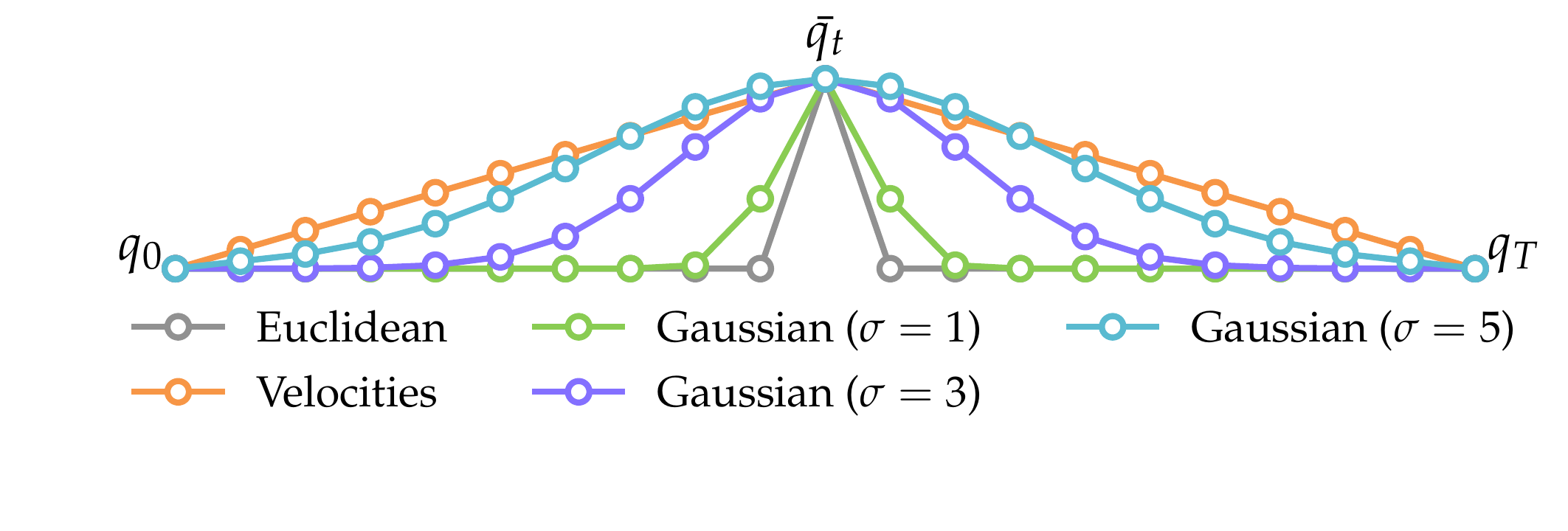}
		\vspace{-10mm}
		\caption{How different norms propagate correction $\bar{q_t}$.}
		\label{fig:curves}
	\end{figure}

Different norms lead to different types of propagations. \figref{fig:curves} shows how different metrics induce different propagation behaviors.

\noindent\textbf{Identity.}
Setting $A$ to be the identity matrix corresponds to using the Euclidean inner product and leads to the optimal solution  $\bar{\xi}=(q_0,..,\bar{q_t},..,q_T)$ as in \cite{jain2013}. However, in this work we hypothesize that different, non-Euclidean inner products perform better at capturing what users intend with their corrections and will lead to faster learning. We present some options below.

\noindent\textbf{Velocities and Higher Order Terms.} One way to induce smoothness is to penalize changes in velocity. If A is the finite differencing matrix:
\begin{equation}
    A[i,j] = \begin{cases}2, &\text{ if $i=j$}\\
    -1, &\text{ if $\lvert i-j\rvert$ = 1}\\
    0, &\text{ else}\end{cases}
\end{equation}
then $\xi\trans A\xi$ computes the sum squared velocities of trajectory $\xi$. Such a matrix and alternatives for higher order derivatives are popular in trajectory optimizers \cite{chomp,trajopt,stomp} to produce smooth trajectories, and have also been used in physical human-robot interaction \cite{losey2017}, including in the context of learning from corrections \cite{bajcsy2017}. 

The inverse of the finite differencing matrix also has an intuitive interpretation, linearly propagating corrections over the entire trajectory (See \figref{fig:curves}).

\noindent\textbf{Gaussian (RBF) Kernel.} 
An alternative is setting $A^{-1}$ to the RBF kernel:
\begin{equation}
    A^{-1}[i, j] \propto \exp{-\frac{(i-j)^2}{2\sigma^2}}
\end{equation}
This provides an additional hyperparameter, $\sigma$, that allows us to tune how local or global the desired propagation is.

Note that when we use an RBF kernel, our estimation update of $\xi$ is analogous to function approximation in Online Kernel Machines \cite{kivinen2004}. There, a new data point adds a new term to the function which propagates the change via the kernel. This is equivalent to Equation \ref{eqn:gradients}, modulo our end point constraints $\xi(0)$ and $\xi(T)$, and the fact that we impose $\xi(t)=\bar{q_t}$ as a hard constraint.

\subsection{Overall Algorithm}

\begin{figure}
		\centering
		\includegraphics[width=0.49\columnwidth]{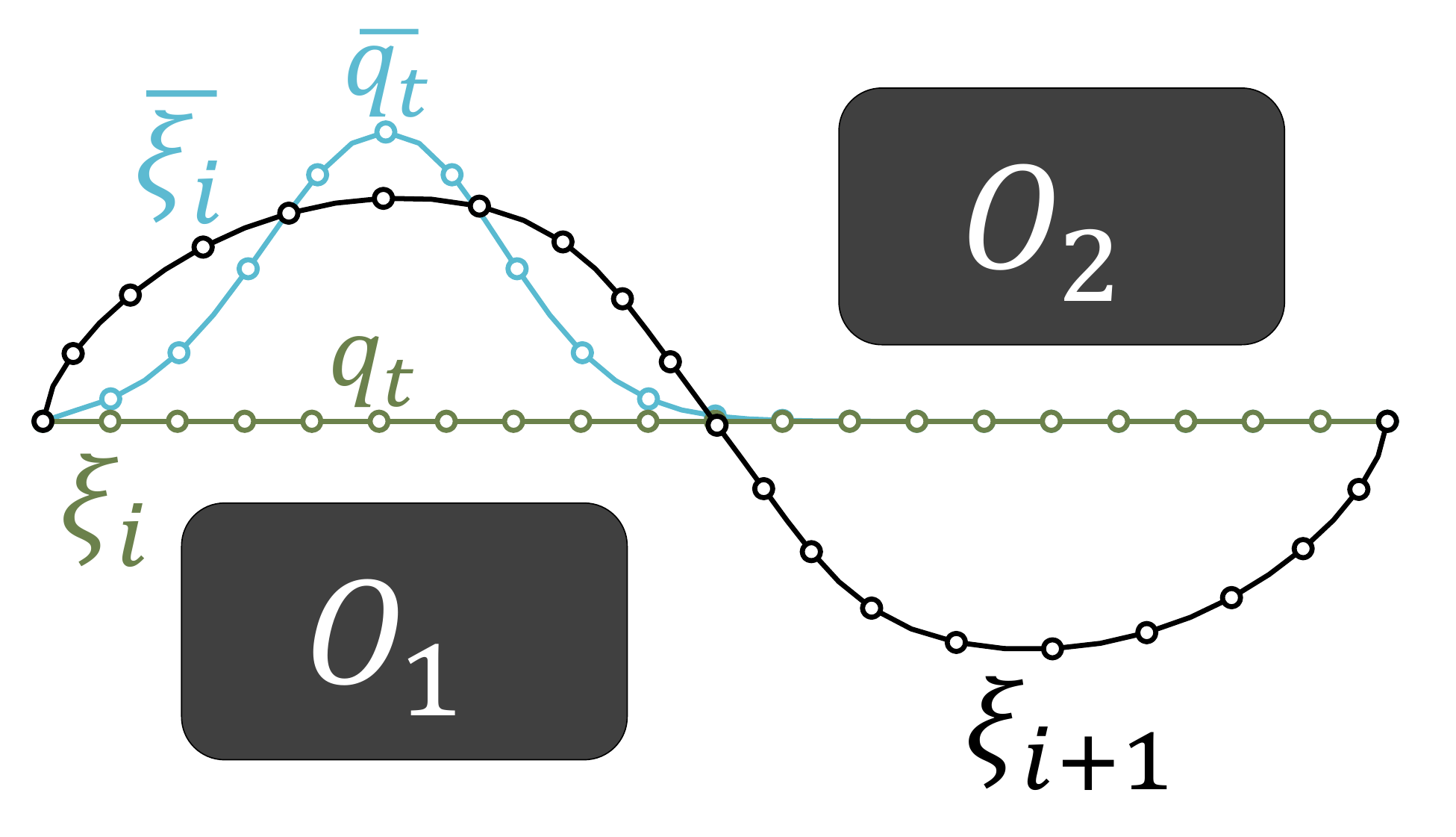}
		\includegraphics[width=0.49\columnwidth]{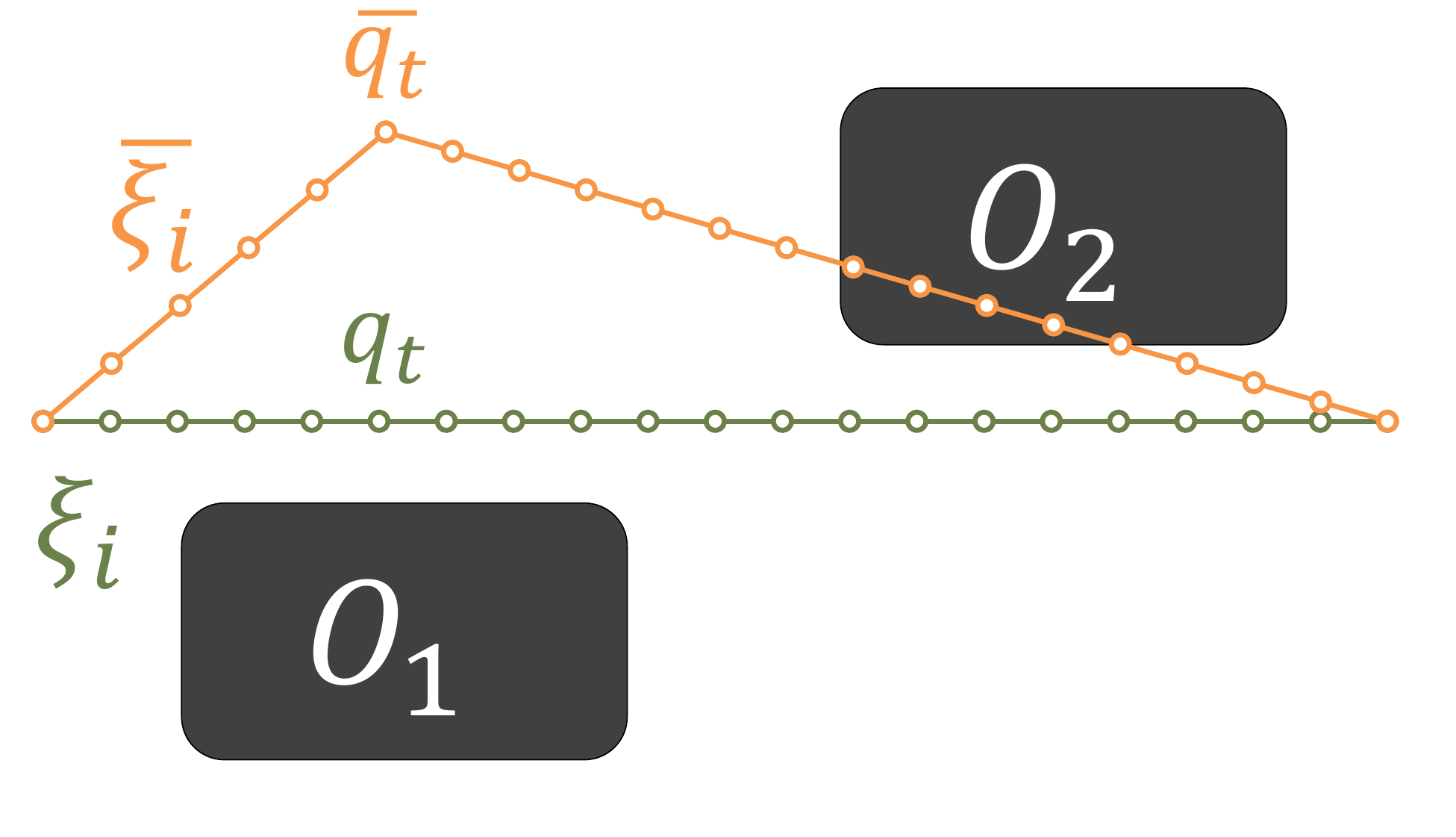}
		\caption{\textbf{Left:} One iteration of the algorithm. User sees current trajectory $\xi_i$ and provides a correction by moving $q_t$ to $\bar{q_t}$ to avoid an obstacle. The robot extrapolates to the trajectory $\bar{\xi_i}$ and updates its weights to produce $\xi_{i+1}$. \textbf{Right:} In some environments, wider propagations can result in the algorithm unintentionally inferring the wrong updates to other features. Thus, the norm that best matches the user's intended trajectory can be task dependent.}
		\label{fig:obstacles}
	\end{figure}

Put altogether, we present the following algorithm (Alg. \ref{alg:lfc}). First, we initialize the weight vector to zero. Each iteration, a trajectory optimizer \cite{trajopt} plans the optimal trajectory $\xi_i$ with current weight vector $w_i$. We present the trajectory to the user who provides a correction $\bar{q_t}$ consisting of a timepoint and a joint configuration. We extrapolate to a full trajectory $\bar{\xi_i}$ by treating the feedback as an online function approximation problem subject to a predefined inner product norm. Finally, we update our weights in the direction of the difference in the features of the trajectory $\phi(\bar{\xi_i})$ and those of the original planned trajectory $\phi(\xi_i)$, weighted by the learning rate $\beta$. \figref{fig:obstacles} depicts one iteration of the algorithm.

Note that as long as the user's corrections produce trajectories with lower cost in expectation, the expected regret of our algorithm has an upper bound of $\mathcal{O}(1/\sqrt{N})$ after $N$ iterations \cite{shivaswamy2015}.

\begin{algorithm}[h!]
	\caption{Learning from (Extrapolated) Corrections}
	\begin{algorithmic}[1]
		\State Initialize $w_0\leftarrow 0$
		\For{$i=1$ to $N$}
		\State $\xi_{i} \leftarrow \arg\min_\xi w_i\trans \phi(\xi)$
		\State Obtain user feedback $\bar{q_t}$
		\State $\bar{\xi_i} \leftarrow \arg\min_\xi \frac{1}{2}||\xi - \xi_i||^2_A$ s.t. $\xi(t) = \bar{q_t}$
		\State $w_{i+1} \leftarrow w_i - \beta (\phi(\bar\xi_i) - \phi(\xi_i))$
		\EndFor		
	\end{algorithmic}
	\label{alg:lfc}
\end{algorithm}

\section{Results in Simulation}

We first provide analysis in simulation to see how the choice of norm impacts performance and whether non-Euclidean norms might be more effective in certain environments.

\begin{figure}[htbp]
\centering
\subfloat[Start and Goal]{\includegraphics[width=.29\columnwidth]{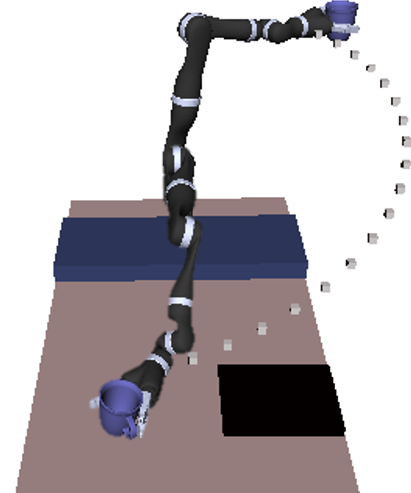}}
\subfloat[Euclidean]{\includegraphics[width=.28\columnwidth]{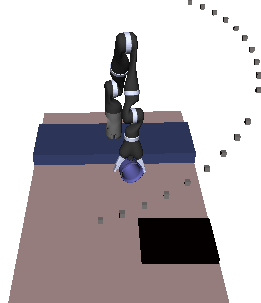}}
\subfloat[Velocities]{\includegraphics[width=.27\columnwidth]{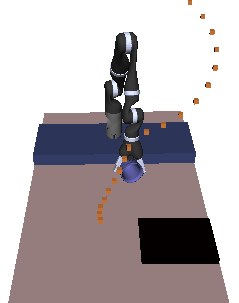}}\\
\subfloat[Gauss. ($\sigma=1$)]{\includegraphics[width=.3\columnwidth]{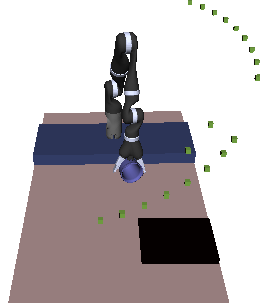}}
\subfloat[Gauss. ($\sigma=3$)]{\includegraphics[width=.28\columnwidth]{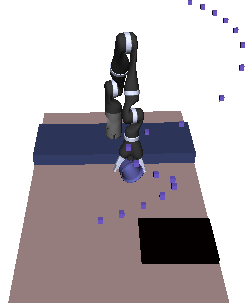}}
\subfloat[Gauss. ($\sigma=5$)]{\includegraphics[width=.28\columnwidth]{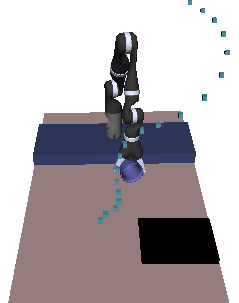}}
\caption{Examples of propagations using different norms for the same correction.}
\label{fig:openrave}
\end{figure}

\begin{figure}[t]
\centering
\includegraphics[width=0.7\columnwidth]{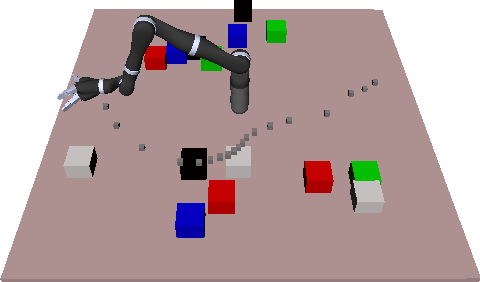}
\caption{Example of a randomly simulated environment with 5 features and 2 instances of each feature. The gray trajectory successfully avoids the green and black boxes (positive weight) while staying close to the blue and red boxes (low weight).}
\label{fig:env}
\end{figure}

\noindent\textbf{Environments.} 
We simulated a set of environments with objects of varying types. We varied the number of such types (features), as well as the number of instances for each type. \figref{fig:env} shows an example environment, and \figref{fig:openrave} shows the effect of each norm on the trajectory. For each environment, we assigned a random ground truth objective function (i.e. weight vector).

\noindent\textbf{Simulating user input.}
For each environment, we use the ground truth weights to plan the ground truth trajectory $\xi^*$. We then iterate by planning a trajectory for the current weights (starting with $w_0$), selecting the timepoint $t$ at which the optimal trajectory and the planned trajectory differ the most (calculated using $\norm{\xi^*(t) - \xi_i(t)}_2^2$), and correcting that waypoint to the its value in the ground truth  trajectory.

\noindent\textbf{Implementation details.}
We used Trajopt \cite{trajopt} as our trajectory optimizer and simulated the environments in OpenRAVE \cite{diankov2010}. For each combination of (number of types, number of instances) pair, we generated 25 different environments. 
To make sure that cost values in different environments were comparable, we normalized costs such that the optimal trajectory had a cost of 0 and the initial trajectory (straight-line in configuration space) had a cost of 1.

In addition, we tuned the learning rate $\beta$ for each norm individually. Since wider kernels tend to update the weight vector more dramatically, we found that the optimal learning rates for such kernels were lower than those of narrower kernels.

\begin{figure}[htbp]
\centering
\includegraphics[width=\columnwidth]{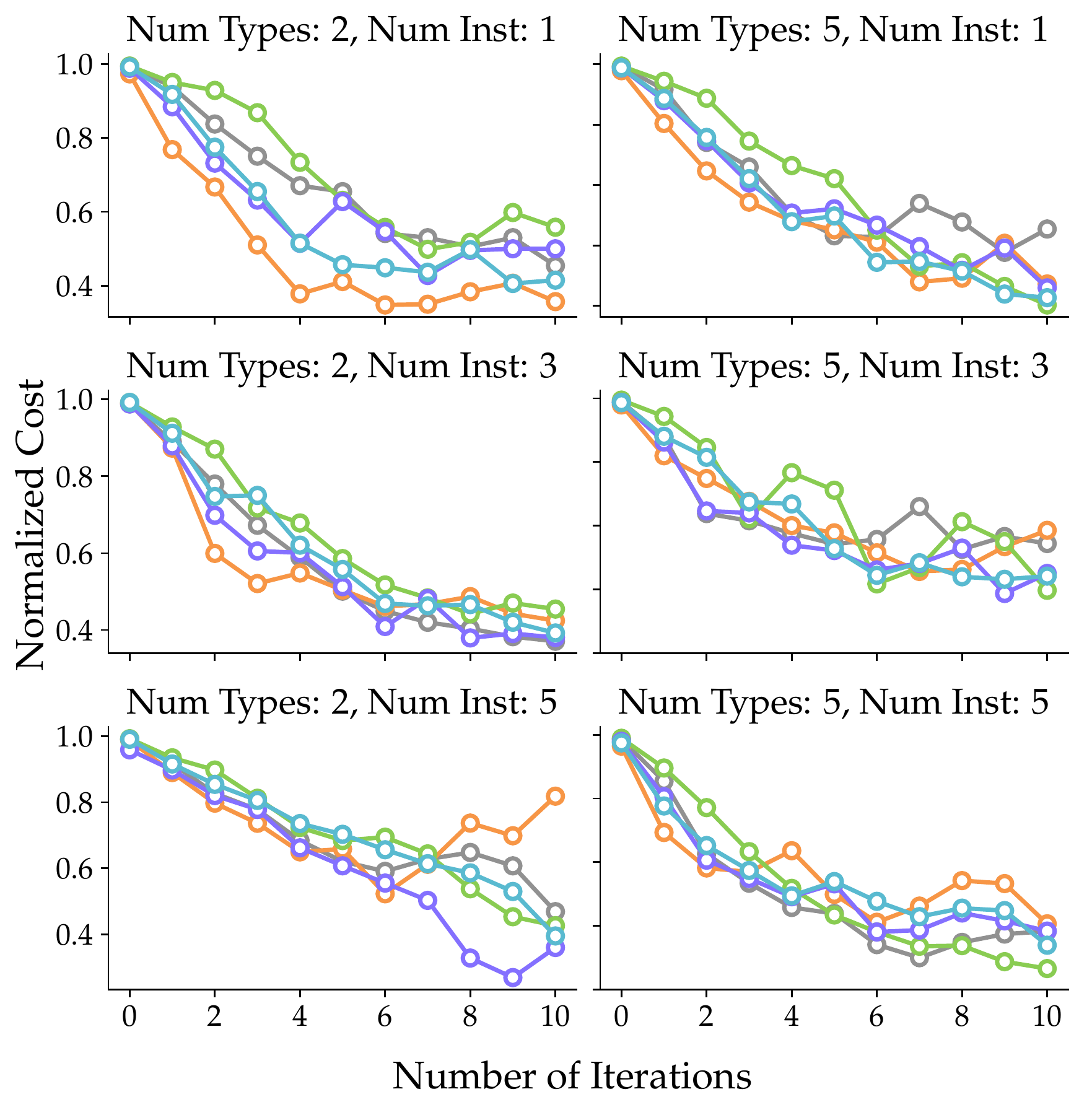}\\
\vspace{-0.5mm}
\includegraphics[width=0.9\columnwidth]{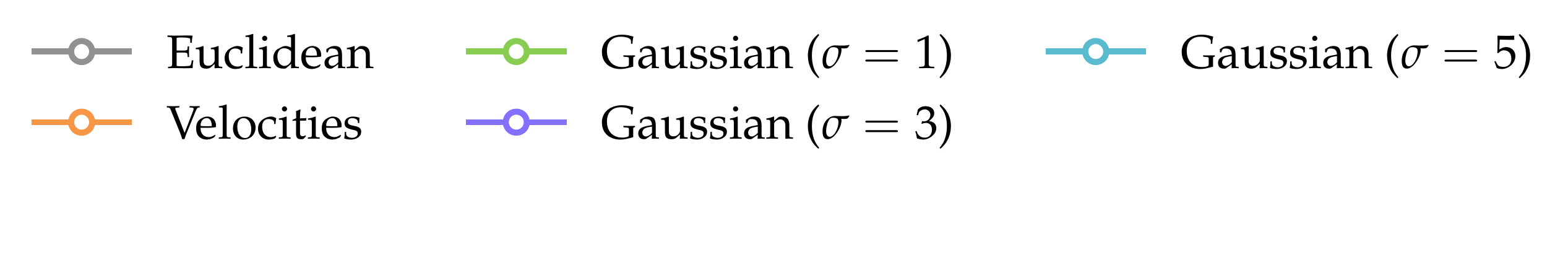}\\
\vspace{-6mm}
\caption{Median cost per iteration in simulation environments using different metrics. We generated 25 different environments for each pair of number of features and feature instances. Learning rates ($\beta$) were tuned per metric.}
\label{fig:sim}
\end{figure}

\noindent\textbf{Analysis.}
\figref{fig:sim} shows the results of the simulations. Overall, we found that in simple environments with few object types and instances, wider propagations (i.e. wider kernels for $A^{-1}$) result in lower cost over time. However, as environments become increasingly complex, norms that produce narrower propagations become more effective as the user has more fine-tuned control over the correction. In environments with many object types, the advantage of wider propagataions is less noticeable. In cluttered environments, wider propagations are more likely to cause the algorithm to unintentionally infer the wrong updates to other features (See \figref{fig:obstacles}). Overall, the velocities norm compares somewhat favorably to the Euclidean distance, with the exception of environments with both high clutter and many feature types. 

\section{User Study}

Our simulations revealed that while there are situations where Euclidean corrections work well, there are also many cases where non-Euclidean is preferable. We designed and conducted a user study to test this result with real users. 

\subsection{Experimental Design}

\noindent\textbf{Task:} We instructed users to teach a JACO2 7-degree-of-freedom arm to plan trajectories that balance three different properties:
\begin{itemize}
    \item keep the cup close to the table,
    \item keep the cup over the table, and
    \item keep the cup away from the laptop.
\end{itemize}
For each iteration, the user provided a correction to the trajectory by selecting one of the waypoints verbally and then physically correcting that waypoint in gravity-compensation mode.

\noindent\textbf{Independent Variables:} We manipulated 3 independent variables: 
\begin{itemize}
    \item Norm for Interpreting Corrections: We tested our algorithm with the Euclidean Norm and the Velocities Norm. The user could provide up to five corrections with each method, stopping if the user felt that the planned trajectory looked exactly like the optimal trajectory. 
    \item Location Strategy: We tested two strategies to come up with user corrections. Users have to decide on the time point $t$ to give the correction $\bar{q_t}$ at, so in one condition, we provided no instructions for selecting the correction and let them choose what intuitively made sense to them (Anywhere); whereas in another condition, we instructed them to choose the time point at which the optimal trajectory and the planned trajectory differed the most (Largest).
    \item Environment: We designed two environments to make sure our results were not specific to a particular setting. We chose the environments such that one might benefit from wider propagations, whereas another might require more local corrections because correcting the different features (distances to table, laptop) might come in conflict. \figref{fig:obstacles} shows an example of how global propagations can lead to counter-productive updates.
\end{itemize}

\noindent\textbf{Dependent Variables:} We designed measures that can capture whether people intend non-Euclidean corrected trajectories, both objectively and subjectively

Since we could not directly access each user's internal preferences, we defined a set of optimal weights and showed the optimal trajectory planned using those weights to the users to ground their preferences and make sure they understood the task. Thus, participants tried to recreate the optimal behavior. This enabled us to measure the robot's performance based on an objective measure of \emph{cost} computed from the optimal weights. If the robot produces trajectories closer to the intended corrections, then those trajectories should have lower cost.

At the end of each iteration, users rated how closely the corrected trajectory matched what they had in mind. This enabled us to measure subjectively how good each norm is at producing the intended correction. We had participants rate the planned trajectory as well to evaluate how well they thought the robot was learning from the correction.
Once they were finished giving corrections using each method, the participants rated how well the robot understood their corrections and the ease of teaching the robot.

\noindent\textbf{Hypotheses:}
    \begin{itemize}
        \item [\textbf{H1:}] The non-Euclidean norm produces trajectories with lower overall cost (by the end of the learning and also along the way).
        \item [\textbf{H2:}] The non-Euclidean norm leads to corrected trajectories that better match what the user intends, as evaluated via self-reports.
         \item [\textbf{H3:}] The user perceives corrections with the non-Euclidean norm as more successful at learning and easier to use for teaching.
    \end{itemize}
    
\noindent\textbf{Participants:} We recruited 26 participants (11M, 15F) from UC Berkeley students. 15 participants reported having a technical background. The norm factor was within-subjects: participants provided corrections interpreted using both norms to provide a calibrated comparison. The location strategy and environment factors were between-subjects: we assigned one environment and one location strategy to each participant. We presented the methods in counterbalanced order.

\subsection{Analysis}

\begin{figure}[ht]
\centering
\includegraphics[width=\columnwidth]{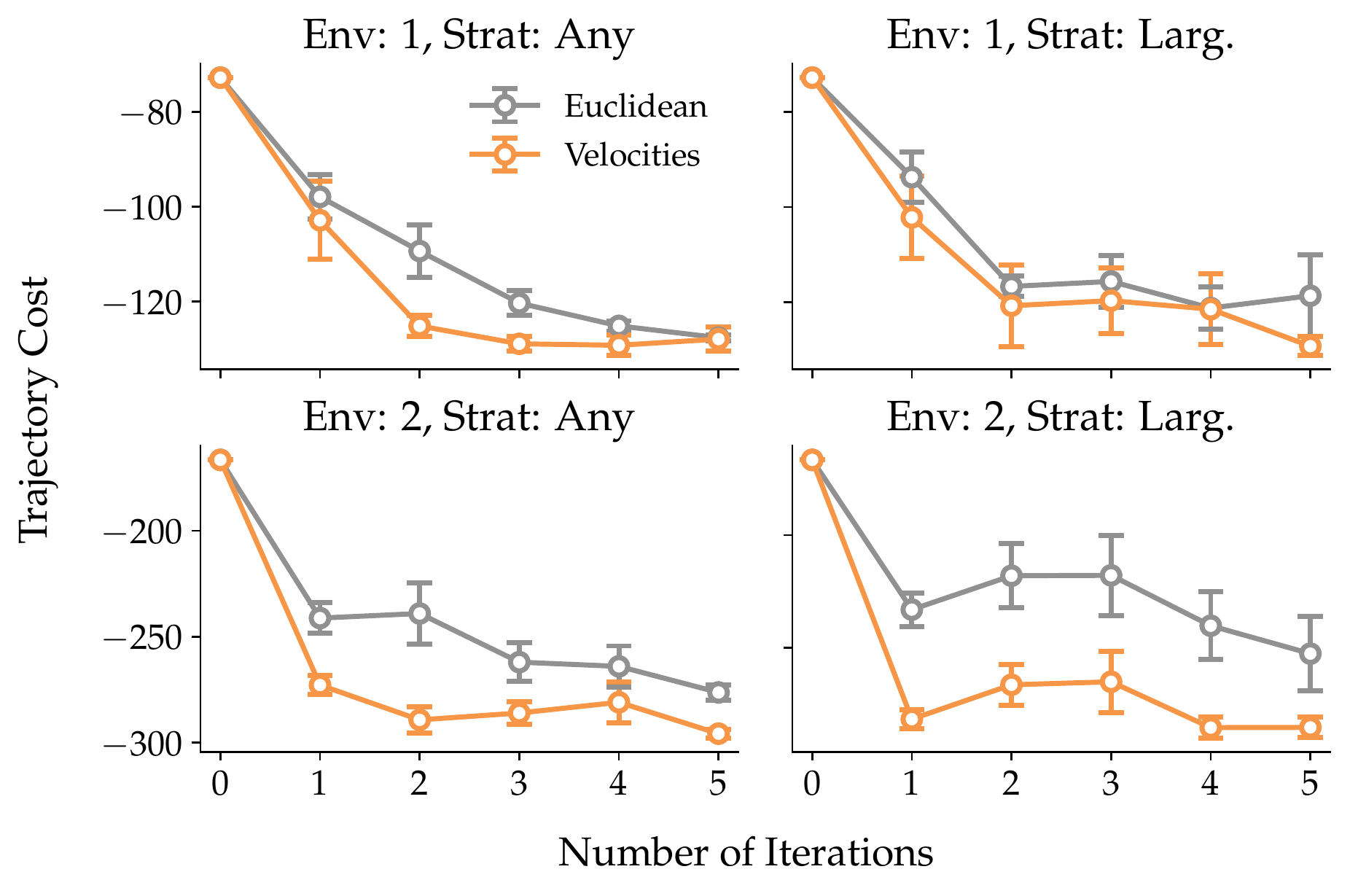}
\caption{Costs per iteration for planned trajectory $\xi_i$ in user study. Error bars show standard error.}
\label{fig:costs}
\end{figure}

\noindent\textbf{Objective:} We conducted a factorial repeated measures ANOVA with environment type (1 or 2), location strategy (anywhere or largest), and norm type (Euclidean or Velocities) as factors, on the cost. We used time (iteration 1 through 5) as a factor as well. We found that environment ($F(1, 22)=1324.916, p<0.0001$), location ($F(1, 22)=4.7184, p=0.0409$), time ($F(5, 277)=91.5669, p<0.0001$), and norm ($F(1, 277)=48.3257, p<0.0001$) have statistically significant effects on cost (See \figref{fig:costs}). 

As expected, cost decreased over time. Surprisingly, the anywhere location strategy was significantly better, suggesting that end-users are intuitively able to choose good points to intervene with a correction.

There was an interaction effect only for environment and norm ($F(1, 277) =24.5420, p<0.0001 $). A post-hoc analysis with Tukey HSD showed that the non-Euclidean norm led to significantly lower cost in environment 2, and lower but only marginally significant in environment 1. Overall, our findings support \textbf{H1}. 

\begin{figure}[t]
\centering
\includegraphics[width=\linewidth]{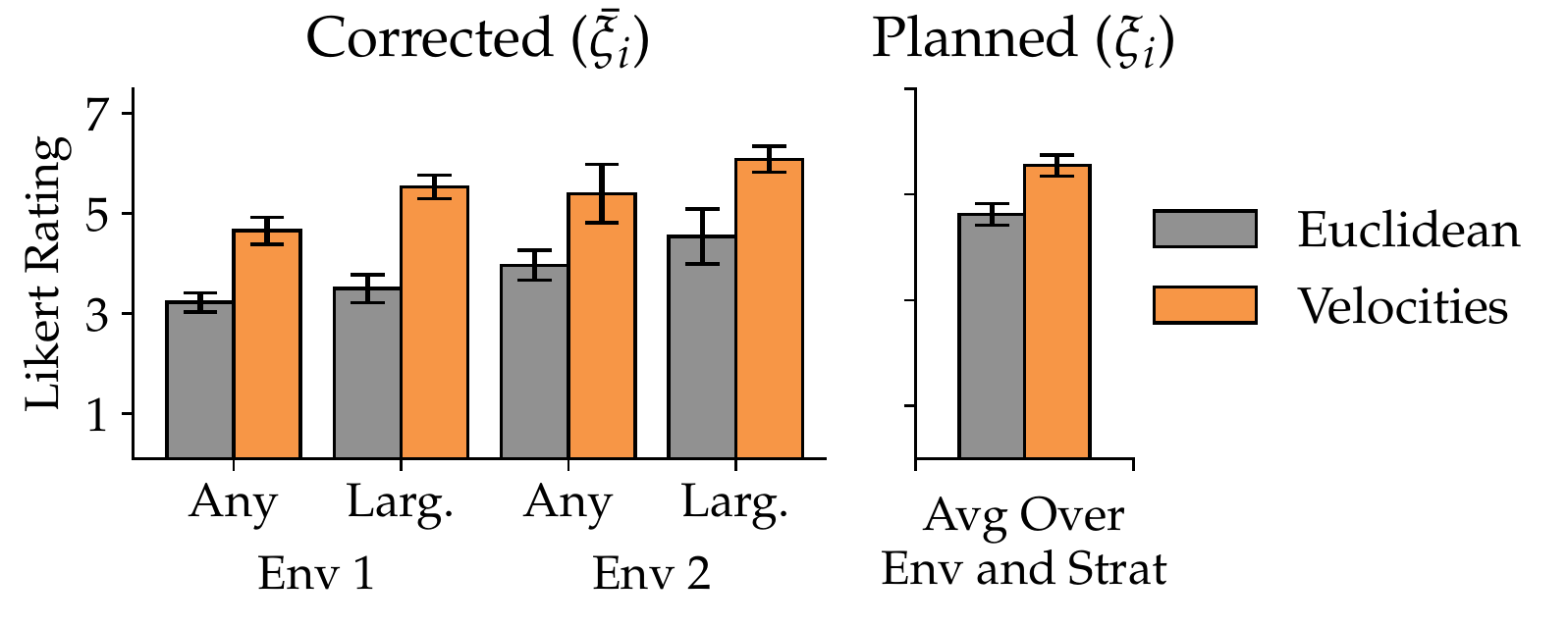}
\vspace{-5mm}
\caption{Each iteration, users rated their agreement with the statements ``The corrected trajectory matches the trajectory that I had in mind" and ``The planned trajectory matches the trajectory that I had in mind."}
\label{fig:ratings}
\end{figure}

\noindent\textbf{Subjective:} We conducted a factorial repeated measure  ANOVA with environment type, location of correction, and norm type as factors on the average rating for corrected and planned trajectories. We found that environment ($F(1, 22)=7.0637, p=0.0144$), location ($F(1, 22)=4.3338, p=0.0492$), and norm ($F(1, 22)=37.1247, p<0.0001$) have a statistically significant effect on corrected trajectory ratings.  No interaction effects were statistically significant. We found that environment 2 was harder for users. Surprisingly, they perceived the anywhere strategy as less effective, even though objectively it performed better. But most importantly, in line with \textbf{H2}, they perceived the non-Euclidean corrections to better match their intended corrections (See \figref{fig:ratings}).

Only norm ($F(1, 22)=11.2232, p=0.0029$) had a statistically significant effect on planned trajectory ratings, in the direction we hypothesized (\textbf{H3}): non-Euclidean corrections led to better planned trajectories.

Finally, we also ran an ANOVA on subjective ratings for the users' experience with each norm. Table \ref{table:statements} summarizes the results. These were also in support of \textbf{H3}.

\begin{table}
    \centering
    \begin{tabular}{lp{5cm}ll}
        \toprule
        & Statement & $F(1,22)$ & p-value  \\\midrule
        1.  & By the end, the robot understood how I wanted it to do the task.
            & 6.6710    & 	\textbf{0.0170}\\
        
        2.  & The robot's performance improved over time.
            & 3.1325    &   0.0906\\
            
        3.  &  I had to keep correcting the robot.
            & 11.2886	&   \textbf{0.0028}\\
        
        4.  &  It was easy to anticipate how the robot would respond to my corrections.
            & 25.2018  &   \textbf{$<$0.0001}\\
        
        5.  & It was easy to physically interact with the robot.
            & 0.0831   &   0.7759\\
        
        6.  & I knew what to do to get the robot to perform the task correctly.
            & 16.0177  &   \textbf{0.0006}\\
        \bottomrule
    \end{tabular}
    \caption{Users rated their agreement with each of these statements after correcting the robot using the Euclidean Norm and the Velocities Norm.}
    \label{table:statements}
\end{table}

\section{Discussion}
\noindent\textbf{Summary.} When receiving a correction, the robot does not observe the entire intended trajectory, instead receiving only a single data point. When we explicitly account for that lack of knowledge, we are faced with an online function approximation problem. Solving it in a non-Euclidean inner product space can lead to better learning in some environments than when making a default assumption about what the user intended, either in lower cost or in fewer required interventions from the user.

\noindent\textbf{Limitations and Future Work. }
The biggest limitation of our method is that it still needs to commit to an inner product or norm in order to interpret the corrections, and while we've found one that worked well for many of the tasks we tested, different tasks might benefit from different norms (See \figref{fig:obstacles}). Future work should investigate ways of learning the desired norm interactively from the user. 

One limitation in the user study is that we present users with an optimal trajectory, thus trading the external validity of having real preferences for the benefit of having a more objective measure of cost. Future work should complement our study with one that seeks users' internal preferences and only evaluates the learning subjectively. 

Further, there could be ways of learning from corrections that do not require the intermediate step of inferring a trajectory. While these might be expensive now (e.g. reasoning about the Q-value of a corrected action rather than the cumulative cost of an entire trajectory), approximation methods for them might entirely bypass the need for an intended trajectory.

\section*{Acknowledgements}

This research was supported by funding from Open Philanthropy, AFOSR, and an NSF Career Award. We thank Andrea Bajcsy for insightful discussion and sharing code. We would also like to thank all members of the members of the InterACT lab for helpful feedback.

\bibliography{arxiv_submission}{}
\bibliographystyle{plain}

\addtolength{\textheight}{10cm}

\end{document}